\documentclass[preprint,12pt]{elsarticle}

\usepackage{amssymb}

\usepackage{lineno}
\usepackage{color}
\usepackage{algorithmic,algorithm}
\usepackage[table]{xcolor}
\usepackage{array}
\usepackage{enumerate}
\usepackage{amsmath}
\usepackage{booktabs}
\usepackage{graphicx}
\usepackage{float}
\usepackage{threeparttable}
\usepackage{epstopdf}
\usepackage{xfrac}
\usepackage{makecell}

\usepackage{subfig}
\usepackage{multirow}

\graphicspath{{./fig/}}

\journal{Information Science}
\begin{document}

\begin{frontmatter}


\title{Evolutionary Multiparty Distance Minimization}


\author[hitsz]{Zeneng She}
\author[hitsz]{Wenjian Luo}
\author[ustc]{Xin Lin}
\author[hitsz]{Yatong Chang}
\author[us]{Yuhui Shi}

\address[hitsz]{School of Computer Science and Technology, Harbin Institute of Technology, Shenzhen 518055, Guangdong, China}
\address[ustc]{School of Computer Science and Technology of the University of Science and Technology of China, Hefei 230027, Anhui, China}
\address[us]{Guangdong Provincial Key Laboratory of Brain-inspired Intelligent Computation, Department of Computer Science and Engineering, Southern University of Science and Technology, Shenzhen 518055, China}
\address{Email: 20s151103@stu.hit.edu.cn, luowenjian@hit.edu.cn, iskcal@mail.ustc.edu.cn, 20s151150@stu.hit.edu.cn, shiyh@sustech.edu.cn}

\begin{abstract}
	In the field of evolutionary multiobjective optimization, the decision maker (DM) concerns conflicting objectives.
	In the real-world applications, there usually exist more than one DM and each DM concerns parts of these objectives.
	Multiparty multiobjective optimization problems (MPMOPs) are proposed to depict the MOP with multiple decision makers involved, where each party concerns about certain some objectives of all.
	However, in the evolutionary computation field, there is not much attention paid on MPMOPs.
	This paper constructs a series of MPMOPs based on distance minimization problems (DMPs), whose Pareto optimal solutions can be vividly visualized.
	To address MPMOPs, the new proposed algorithm OptMPNDS3 uses the multiparty initializing method to initialize the population and takes JADE2 operator to generate the offsprings.
	OptMPNDS3 is compared with OptAll, OptMPNDS and OptMPNDS2 on the problem suite.
	The result shows that OptMPNDS3 is strongly comparable to other algorithms.
\end{abstract}


\end{frontmatter}


\section{Introduction}\label{sec:introduction}

In real world, decision-making problems tend to have multiple conflicting objectives, where decision makers (DMs) often require the best solutions.
Such decision-making problems with two or three objectives are called multiobjective optmization problems (MOPs) \cite{coello2006evolutionary,fonseca1995overview,tamaki1996multi} and those with more than three objectives are named as many-objective optimization problems (MaOPs) \cite{ishibuchi2008evolutionary,von2014survey,li2015many}.
MOPs are import problems and play significant roles in engineering application including industrial design \cite{rosso2020multi,lin2021multi}, scheduling \cite{ming2017deriving,chiang2019multi}, project portfolio optimization \cite{goli2019hybrid,silva2019multi} and vehicle routing \cite{zhang2019hybrid,braekers2016vehicle}.
The MOP attracts interest of many researchers from various domains including mathematicians, computer scientists, engineers and economists for its applications in different fields \cite{chand2015evolutionary}.

Since the objectives of MOPs are conflicting, optimizing one objective may lead to deterioration of others in MOPs.
Under the situation, Pareto optimal solutions, which no solution is better than in all objectives, are answers to the problems.
Given no further information, Pareto optimal solutions are offered to the decision maker to choose.
To obtain Pareto optimal solutions, evolutionary methods are widely used for they can generate a host of solutions in one run without further information.
A great many of MOEAs hava been proposed such as: NSGA-II \cite{deb2002fast}, NSGA-III \cite{deb2013evolutionary}, PEAS-II \cite{corne2001pesa}, SPEA2 \cite{zitzler2001spea2}, MOEA/D \cite{zhang2007moea}, Two\_Arch2 \cite{wang2014two_arch2}, and HypE \cite{bader2011hype}.

However, there is usually more than one decision maker involved in the real world applications.
These decision maker usually does not care about all objectives in MOPs but a part.
Multiparty multiobjective optimization problems (MPMOPs) are a special class of MOPs, where involves multiple parties and each party stands for a DM.
In MPMOPs, there is more than one party and at least one party has more than one objective.
Using MOEAs to handle with MPMOPs will obtain a series of solutions which are not Pareto optimal in the view of all parties.
For example, the article \cite{Ishibuchi2011} introduces a real-world distance minimization problem (DMP), which is easily visualized for its 2-dimensional decision space.
The problem is to find an optimal position which is closest to a set of predefined target points in an actual map.
Four objectives to be minimized in the problem as follows.
\begin{itemize}
	\item $f_1(x)$: Distance to the nearest elementary school,
	\item $f_2(x)$: Distance to the nearest junior-high school,
	\item $f_3(x)$: Distance to the nearest convenience store,
	\item $f_4(x)$: Distance to the nearest railway station.
\end{itemize}
Suppose it is a living position problem about how to locate the best position for a family to purchase or rent their house, treating the house-buying problem as an MOP is not appropriate.
In fact, it is not necessary that everyone in a family should focus on all the objectives.
The positions of elementary and junior-high schools ($f_1(x)$ and $f_2(x)$) generally are only focused by the student in the family.
The parent maybe cares more about the positions of the convenience store and the railway station ($f_3(x)$ and $f_4(x)$).
There are two decision makers, which focus on different objectives, in the living position problem.
Given two solutions $x$ and $y$, $x$ and $y$ are not dominated by each other in the view of the student, i.e., $f_1(x) \leq f_1(y), f_2(x) \geq f_2(y)$; $x$ dominates $y$ in the view of the parent, i.e.,$f_3(x) < f_3(y), f_4(x) \leq f_4(y)$.
As a result, $x$ and $y$ do not dominate each other by comparing the all objectives.
In fact, $x$ is obviously better than $y$ for they are equal good for the student and $x$ is better for the parent.

There is only a little work focusing on the MPMOPs.
In \cite{liu2020evolutionary}, Liu \textit{et al.}  proposed a multiparty algorithm called OptMPNDS based on NSGA-II \cite{deb2002fast} to address the MPMOPs with common Pareto optimal solutions, which are Pareto optimal for all parties simultaneously.
In 2021, She \textit{et al.} \cite{she2021new} proposed a modified algorithm based on OptMPNDS, namely OptMPNDS2, which defines a new dominance relationship under the multiple parties.
However, as an emerging topic in the computational intelligence field, there are still many aspects to be further studied.
Yet, the study about the MPMOP mainly concentrates on the design of algorithms and there is no work about the real-world applications of the MPMOPs.
Consequently, this paper studies the MPMOP from the viewpoint of applications, which applies distance minimization problems (DMPs).
The DMP is a visual multiobjective optimization problem.
So the adaptation of DMPs helps to learn the behaviors of the multiparty multiobjective optimization algorithm.
The primary contributions of this paper are as follows:
\begin{itemize}
	\item 	We constructed a multiparty multiobjective optimization problem suite which is based on distance minimization problems. It is the first work about the applications of MPMOPs. This application can visualize the behaviors of algorithms in decision space.
	\item 	A multiparty multiobjective algorithm, namely OptMPNDS3, is proposed. The algorithm uses the Pareto optimal solutions of every DM as initial population and JADE2 operator to generate offspring. OptMPNDS, OptMPNDS2 and OptMPNDS3 are compared together on the problem suite.
	      Experimental results show that OptMPNDS3 has a comparative performance on the problem suite.
\end{itemize}

The rest of his paper is organized as follows.
First, in Section \ref{sec:related-work}, we introduce the related concepts of the multiobjective optimization problem, the multiparty multiobjective optimization problem, and the distance minimization problem.
We construct new multiparty multiobjective optimization test problems based on several distance minimization problems in Section \ref{sec:the-multiparty-multiobjective-test-problem}.
Next, we propose an evolutionary algorithm for the problem suite in Section \ref{sec:proposed-algorithm}.
Then, Section \ref{sec:experiment} gives the experimental results of four algorithms on the test problems.
Finally, we make a brief conclusion in Section \ref{sec:conclusion}.

\section{Related Work}\label{sec:related-work}
\subsection{Multiobjective Optimization Problems}

Multiobjective optimization problems (MOPs) are a type of optimization problems which have two or three conflicting objectives.
And many-objective optimization problems (MaOPs) have more than three conflicting objectives.
These conflicting objectives cannot be optimal simultaneously.
An MOP or MaOP is defined as a minimization problem without loss of generality \cite{infeasible}:

\begin{equation}\label{eq:MOP}
	\begin{aligned}
		 & Minimize \quad F(x)=(f_1(x),f_2(x),\dots,f_m(x)), \\
		 & Subject \quad to
		\begin{cases}
			h_i(x)=0,\quad i=1,\dots,n_p     \\
			g_j(x)\leq 0,\quad j=1,\dots,n_q \\
			x \in [x_{min},x_{max}]
		\end{cases}
		,
	\end{aligned}
\end{equation}
where $x$ represents a vector of decision variables, of which lower and upper bounds are $x_{min}$ and $x_{max}$, respectively.
$h_i(x) = 0$ denotes the $i$-th equality constraint of $n_p$; $g_j(x) \leq 0$ denotes the $j$-th inequality constraint of $n_q$.
$f_n(x)$ stands for objectives of the $n$-th objectives, and the number of objectives is $m$.

Since the objectives of an MOP are conflicting with each other, the optimizer of an MOP obtains the Pareto optimal solutions according to the Pareto dominance.
Pareto dominance is a binary relation about decision vectors \cite{zhou2011multiobjective}.
Given two decision vectors $x$ and $y$, $x$ dominates $y$, denoted as $x \prec y$, if and only if $f(x) \leq f(y)$ for all objectives and $f_n(x) < f_n(y)$ for at least one objective $f_n$ \cite{nonlinear}.
Based on the dominance relation, a Pareto optimal solution $x$, namely a non-dominated solution, is not dominated by any solutions in the whole decision space $\Omega$, i.e., $\nexists x' \in \Omega, x' \prec x$.
The Pareto optimal set (PS) is a set consisting of Pareto optimal solutions, which is formally defined as PS $= \{ x \in \Omega \vert \nexists x' \in \Omega, x' \prec x \}$.
The Pareto optimal front (PF) is a set of objectives of PS, formally defined as $\{f = (f_1(x),f_2(x),\cdots,f_m(x))| x \in PS\}$.

\subsection{Multiparty Multiobjective Optimization Problems}
A multiparty multiobjective optimization problem (MPMOP) is a special kind of MOPs, which involves multiple parties.
Each party stands for a decision maker and focuses on a part of objectives.
In an MPMOP, the objective function of Formula \eqref{eq:MOP} becomes the following form:
\begin{equation}
	\begin{aligned}
		 & F(x) = (f_1(x),f_2(x),\dots,f_M(x)),      \\
		 & f_i(x) = (f_{i1},f_{i2}, \dots, f_{ii_m})
	\end{aligned}
\end{equation}
where $M$ stands for the number of parties;
$i_m$ represents the quantity of objectives of $i$-th party.

Using the normal multiobjective optimization evolutionary algorithms will obtain a host of solutions which are not optimal for multiple parties.
If only two solutions are not Pareto dominated each other in one party, they are not Pareto dominated each other in all objectives.
Only when a solution is better than another in one party and equal good or better in the rest parties, it is said that the solution is better.

Thus, Liu \textit{et al.} \cite{liu2020evolutionary} proposed an algorithm based on NSGA-II, called OptMPNDS, to address the MPMOPs with common Pareto solutions, which are optimal in each MOP of each party.
To demonstrate the strength of OptMPNDS, the authors also showed the baseline algorithm, namely OptAll, which uses NSGA-II to optimize MPMOPs but the dominated solutions for multiparty are eliminated from the population in the final generation.
Then, She \textit{et al.} \cite{she2021new} proposed a modified algorithm based on OptMPNDS, namely OptMPNDS2, which could obtain better performance on some questions of the benchmark.
Both two algorithms redefine the dominance relation of solutions according to the Pareto levels of all parties, which are obtained from the non-dominated sorting in NSGA-II.
The main difference among the three algorithms including NSGA-II is the sorting of individuals.
As for OptMPNDS, first, it applies the non-dominated sorting to each party to obtain multiple levels for each individual.
Then, the common solutions, whose levels of all parties are the same, are assigned with the same level as the number; the rest solutions are assigned with the max level of all parties as the number.
All the solutions are sorted into different ranks according to the numbers in ascending order.
When the numbers of the former and the latter are the same, the former are prior and placed in front of the latter.
Final, in each rank, crowding distance is used to sort individuals with the same rank.
While, after obtaining the multiple levels of all parties, OptMPNDS2 takes all the levels of each individual as its ``objectives” to perform non-dominated sorting again.

\subsection{Distance Minimization Problems} \label{subsec:Distance Minimizaton Problems}

To evaluate solutions obtained by MOEAs, many indicators are proposed such as Generational Distance (GD) \cite{van2000measuring}, Inverted Generational Distance (IGD) \cite{zhang2008multiobjective} and Hypervolume (HV) \cite{zitzler1999multiobjective}.
However, DMs usually do not understand the indicators for lack of professional knowledge.
And some MOEAs get into trouble when facing the many-objective optimization problems.
Consequently, the distance minimization problem (DMP) is proposed to study the behaviors of MOEAs recently since its 2-dimensional or 3-dimensional decision space is easily visualized.

The DMP is to find an optimal position which is closest to a set of predefined target points in 2-dimensional or 3-dimensional space. 
The DMP is a classic MOP because minimizing the distances to multiple target points simultaneously is required in order to solve the problem.
Koppen \textit{et al.} \cite{Koppen2007} constructed the problem that, given a set of points, the objective function is the Euclidean distances of a point to the predefined points. They proved that the Pareto optimal set of the problem equals the convex closure of the set of points.
Rudolph \textit{et al.} \cite{Rudolph2007} introduced a variant of the DMPs, where the equivalent Pareto subsets are distributed in multiple disconnected regions instead of single polygon.
Ishibuchi \textit{et al.} \cite{Ishibuchi2010} defined the general form of the variant that each objective is not the distance to one point, but to the nearest point of a set points.
And they used the DMPs to test MOEAs and generate a real-world many-objective test problem from a real map \cite{Ishibuchi2011}.
Later, they further generated the DMPs from 2-dimensional decision space to $n$-dimensional decision space and tested performances of algorithms by the DMPs with 10, 100 and 1000 dimensions \cite{Ishibuchi2013}.
Zille \textit{et al.} \cite{Zille2015} modified the DMP by using the Manhattan distance as the objective function to substitute the Euclidean distance and found it more difficult to address.
And they generated the dynamic DMPs by reshaping the decision space, changing the distance measure, and so on.
Xu \textit{et al.} \cite{Xu2015} proposed a fast procedure to construct the Pareto optimal set of DMPs under the Manhattan distance measure and proved the optimality of the theoretical construction process.

\subsection{JADE2}
Differential evolution (DE) is an evolutionary algorithm which uses the difference of population individuals to direct the evolution.
The procedure of DE is initializing population and then repeating mutation, crossover and selection operators until the terminated condition is satisfied.

JADE2 \cite{Zhang2008} is a self-adaptive DE algorithm \cite{tanabe2013success,Zuo2021} used to solve multiobjective optimization problems based on JADE \cite{Zhang2009}, which controls the parameters $CR$ and $F$ automatically.
It generates the two parameters for every individual $x_i$, which are chosen from two random distributions respectively.
\begin{equation}
	\begin{aligned}
		CR_i & = randn_i(\mu_{CR}, 0.1) \\
		F_i  & = randc_i(\mu_F, 0.1)
	\end{aligned}
\end{equation}
where $randn_i(\mu_{CR}, 0.1)$ is the number randomly selected from normal distribution with mean $\mu_{CR}$ and variance $0.1$; $randc_i(\mu_F, 0.1)$ is the number randomly selected from Cauchy distribution with local parameter $\mu_F$ and scale parameter $0.1$.
If the obtained value for $CR_i$ is out of [0, 1], $CR_i$ is truncated to 0 or 1.
If the obtained value for $F_i$ is greater than 1, $F_i$ is truncated to 1;
If $F_i$ is less than 0, a value is selected again from the distribution.

$\mu_{CR}$ and $\mu_F$ are updated as follows for each generation, which are initialized to 0.5.
\begin{equation}
	\begin{aligned}
		\mu_{CR} & = (1-c) \cdot \mu_{CR} + c \cdot mean_A(S_{CR}) \\
		\mu_F    & = (1-c) \cdot \mu_F + c \cdot mean_L(S_F)
	\end{aligned}
\end{equation}
where $c$ is a learning rate which is set to 0.1.
$S_{CR}$ and $S_F$ are two collections storied the successful $CR$ and $F$ of previous generations, which are chosen from the corresponding individual when the trial vector generated is better than it corresponding parent.
$mean_A(.)$ is arithmetic mean and $mean_L(.)$ is calculated as follows:
\begin{equation}
	mean_L(S_F) = \frac{\sum_{F \in S_F}F^2}{\sum_{F \in S_F} F}
\end{equation}

Besides, JADE2 uses `DE/current-to-pbest/1' as the mutation operator and an external archive to maintain the diversity.
The operator `DE/current-to-pbest/1' chooses pbest $x_{Pbest}$ randomly from top $p$ percent individuals instead of the best individual.
\begin{equation} \label{eq:DE}
	\begin{aligned}
		u_i(t) = x_i(t) & + F * (x_{Pbest}(t) - x_i(t)) \\
		                & + F * (x_{i1}(t) - x_{i2}(t))
	\end{aligned}
\end{equation}
where $x_{i2}$ is randomly chosen from the population;
$x_{i1}$ is randomly chosen from the union of the population and the archive to help algorithm to explore the bigger region to maintain diversity.
The external archive is used to store individuals which are not selected.
After updating all individuals in a generation, if the archive exceeds the predefined size, some individuals are randomly eliminated until satisfying the size limit.

To address MOPs, the selection of JADE2 includes two stages.
\begin{enumerate}[(1)]
	\item When finishing generating trial vectors, the trial vector is compared with its parent by dominance relation.
	      The non-dominated solution is picked to next stage; the dominated solution is added to the archive.
	\item The left solutions of the first stage participate the non-dominance sorting which NSGA-II uses. And the better ones survive; The worse ones are added to the archive. 
\end{enumerate}

\section{The Multiparty DMPs}\label{sec:the-multiparty-multiobjective-test-problem}

In \cite{Ishibuchi2010}, Ishibuchi \textit{at el.} demonstrates the use of 2-dimensional distance minimization problems (DMPs).
It aims to obtain a point in a 2-dimensional region whose distances to sets of given points should be minimized simultaneously.
The general form of the DMP, namely an unconstrained MOP, is as follows:
\begin{equation}\label{eq2}
	Minimize\ f(x)\ =\ (f_1(x),\ f_2(x),\dots,f_m(x)),
\end{equation}
where $x$ represents the 2-dimensional decision vector.
$f_n$ denotes the $n$-th objective of $m$, which is calculated by the Euclidean distance to a set of $k$ points $(p_{n1}, p_{n2},\dots, p_{nk})$.
Specifically, the distance to the nearest point of the set is practically used as $f_n(x)$:
\begin{equation}\label{eq3}
	f_n(x)= \min \{dis(x,\ p_{n1}),\ dis(x,\ p_{n2}),\ \cdots,\ dis(x,\ p_{nk})\},
\end{equation}
where $dis(x, p)$ means the distance between points $x$ and $p$.

To construct a DMP with $m$ objectives, $m$ sets of points are required, and a set has $k$ points.
Moreover, problems are not identical because of using different $k$.
Based on Formulas \eqref{eq2} and \eqref{eq3}, Ishibuchi \textit{et al.} uses the DMP to examine evolutionary algorithms including NSGA-II in the occasion $k= 1$ and 2.
For $k=1$, the Formula \eqref{eq3} becomes the form as follows.
\begin{equation} \label{eq:dis}
	f_n(x) = dis(x, p_{n1})
\end{equation}

The multiparty distance minimization problem (MPDMP) can be constructed by adding a party to each target point.
The multiparty multiobjective optimization problems with two parties, each of which has two objectives, based on the DMPs are defined as follows:
\begin{equation}\label{eq4}
	\begin{aligned}
		 & Minimize \quad F = (F_1, F_2)                                                \\
		 & F_1 = (f_1, f_2, \dots, f_{i}), \quad F_2 = (f_{i+1}, f_{i+2}, \dots, f_{j})
	\end{aligned}
\end{equation}
where $F_1$ represents the objective function of the party who cares about from $f_{1}$ to $f_{i}$,  $F_2$ represents the objective function of another party who cares about the rest objectives i.e., from $f_{i+1}$ to $f_{j}$.
$f_n(x)$, which is the same as Formula \eqref{eq:dis}~(an objective of a DMP), represents the distance to the predefined target point.

\begin{figure}
	\centering
	\includegraphics[width=.8\textwidth]{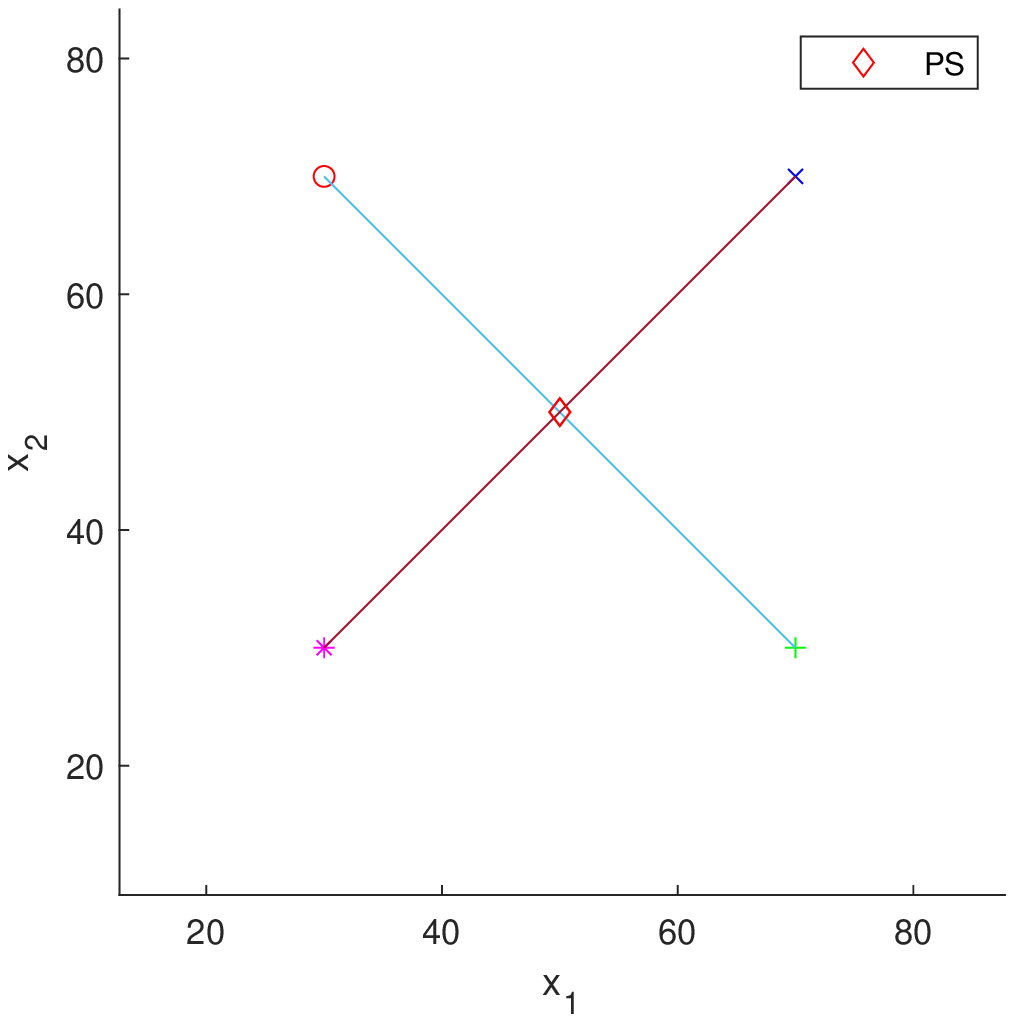}
	\caption{MPDMP1, each line represents the PS of one party}
	\label{figMPDMP1}
\end{figure}

\begin{figure}
	\centering
	\includegraphics[width=.8\textwidth]{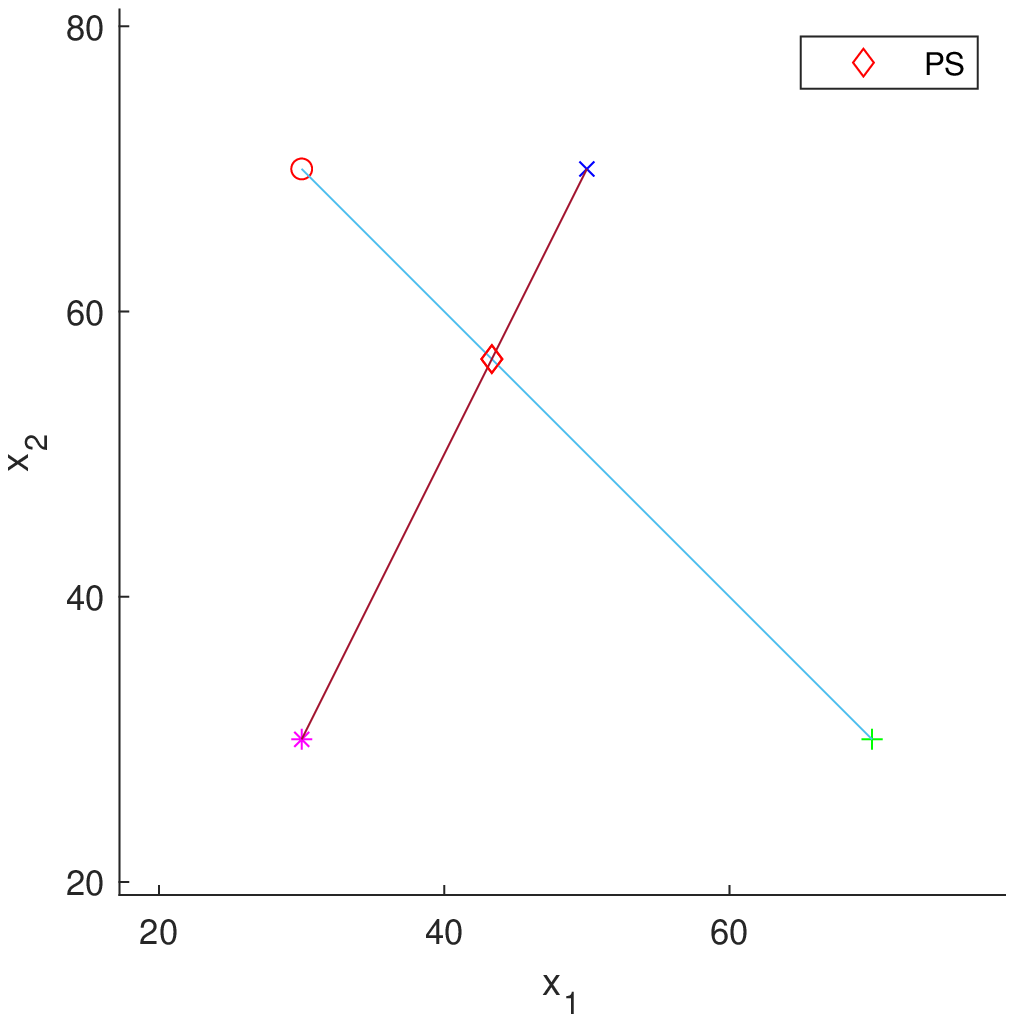}
	\caption{MPDMP2, each line represents the PS of one party}
	\label{figMPDMP2}
\end{figure}

The test problem suite in this paper is constructed according to the Formula \eqref{eq4}.
There are totally 8 problems in the test suite, which are described as follows.
They are constructed from two overlapped geometric shapes, which belong to two parties respectively.
These problems constructed through the above method have common Pareto solutions.
The vertexes of each geometric shape are the target points of one party and the Euclidean distances to the vertexes are objectives of the party.
The first two problems are generated from four target points, where two points are for each party.
In view of each party, the DM faces an MOP with two objectives which are the distances to above two points.
As Figures \ref{figMPDMP1} and \ref{figMPDMP2} shown, for each party, the line connected two target points is the region of his or her Pareto solutions.
And the Pareto solution for all parties is the intersection of each MOP of each party.
Consequently, the PS of MPDMP1, as the same as MPDMP2, is the tagged intersection point of two lines.
The target points of MPDMP3 are the vertexes of a line and a triangle, where each shape for one party.
As it proved in \cite{Koppen2007}, the convex closure of target points is the Pareto solutions in a DMP.
The line is the PS of one party and the triangle is the PS for another party.
Consequently, the PS of MPDMP3 is the intersection region of the line and triangle.

MPDMP4 and MPDMP5 are constructed from two overlapped regular triangles, of which the difference is the region of the PS.
The PS of MPDMP4 is one point and one triangle for MPDMP5.

Similarly, MPDMP6 and MPDMP7, where PSs are a line and a rectangle, respectively, are constructed from two overlapped rectangles.
MPDMP8 is constructed from two overlapped regular pentagons, of which the PS region is a pentagon.
\begin{table}
    \centering
    \caption{The shapes of PS for all problems}
    \label{tab1}
    \resizebox{\textwidth}{!}{
        \begin{tabular}{llll}
            \hline
            Problem & PS of Party 1 & PS of Party 2 & PS            \\
            \hline
            MPDMP1  & one line          & one line          & one point     \\
            MPDMP2  & one line          & one line          & one point     \\
            MPDMP3  & one line          & one triangle      & one line      \\
            MPDMP4  & one triangle      & one triangle      & one point     \\
            MPDMP5  & one triangle      & one triangle      & one triangle  \\
            MPDMP6  & one rectangle     & one rectangle     & one line      \\
            MPDMP7  & one rectangle     & one rectangle     & one rectangle \\
            MPDMP8  & one pentagon      & one pentagon      & one pentagon  \\

            \hline
        \end{tabular}
    }
\end{table}

\begin{figure}
	\centering
	\includegraphics[width=.8\textwidth]{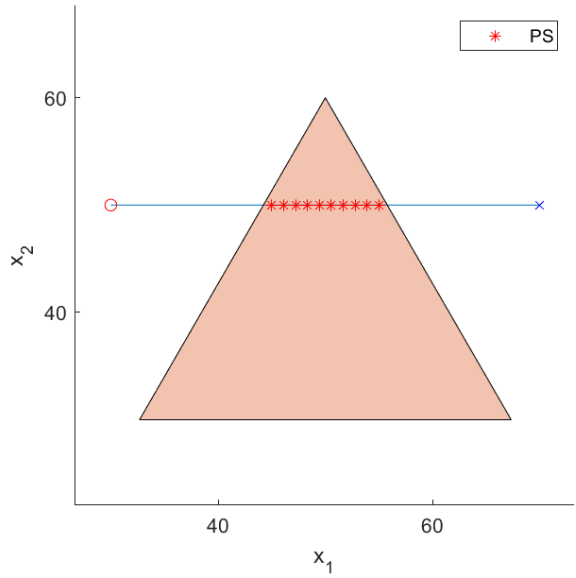}
	\caption{MPDMP3, the line represents the PS of one party and the triangle for another}
	\label{figMPDMP3}
\end{figure}

\begin{figure}
	\centering
	\includegraphics[width=.8\textwidth]{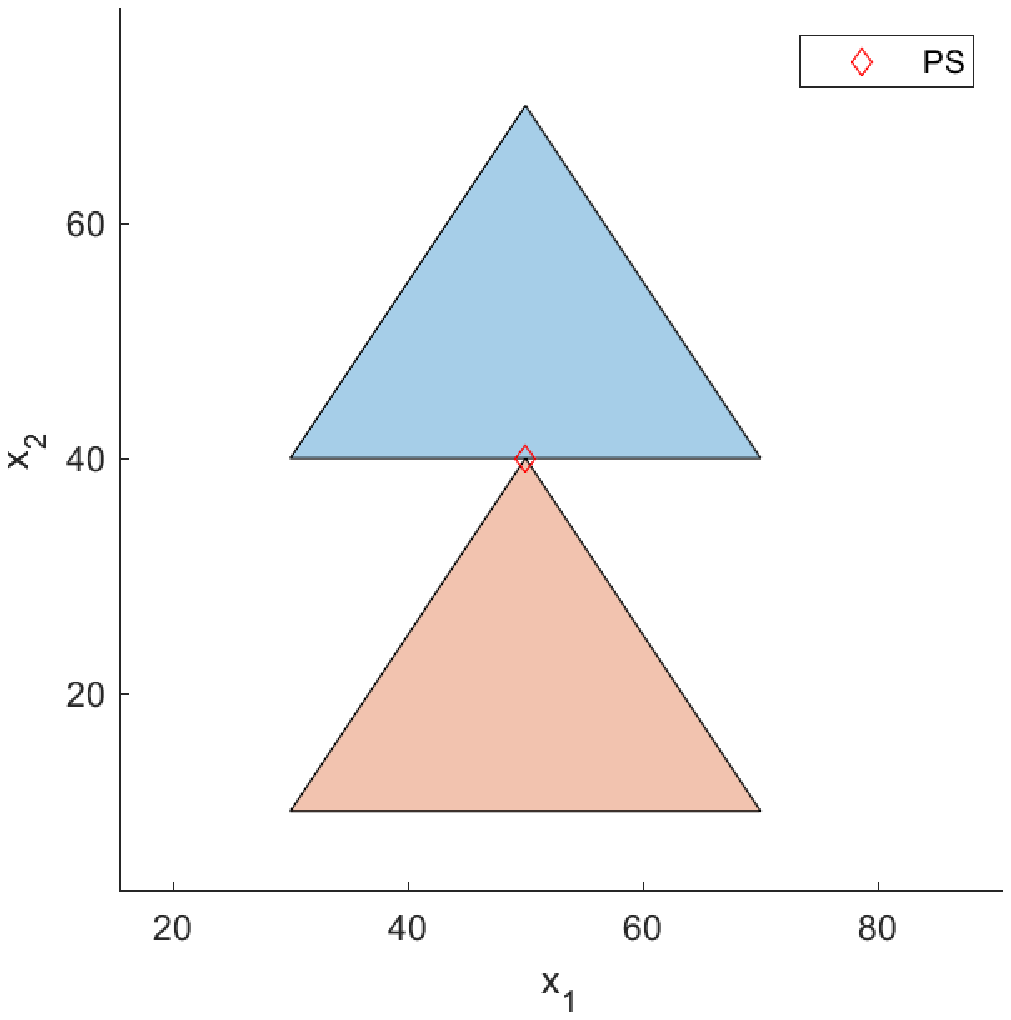}
	\caption{MPDMP4, each triangle represents the PS of one party}
	\label{figMPDMP4}
\end{figure}

\begin{figure}
	\centering
	\includegraphics[width=.8\textwidth]{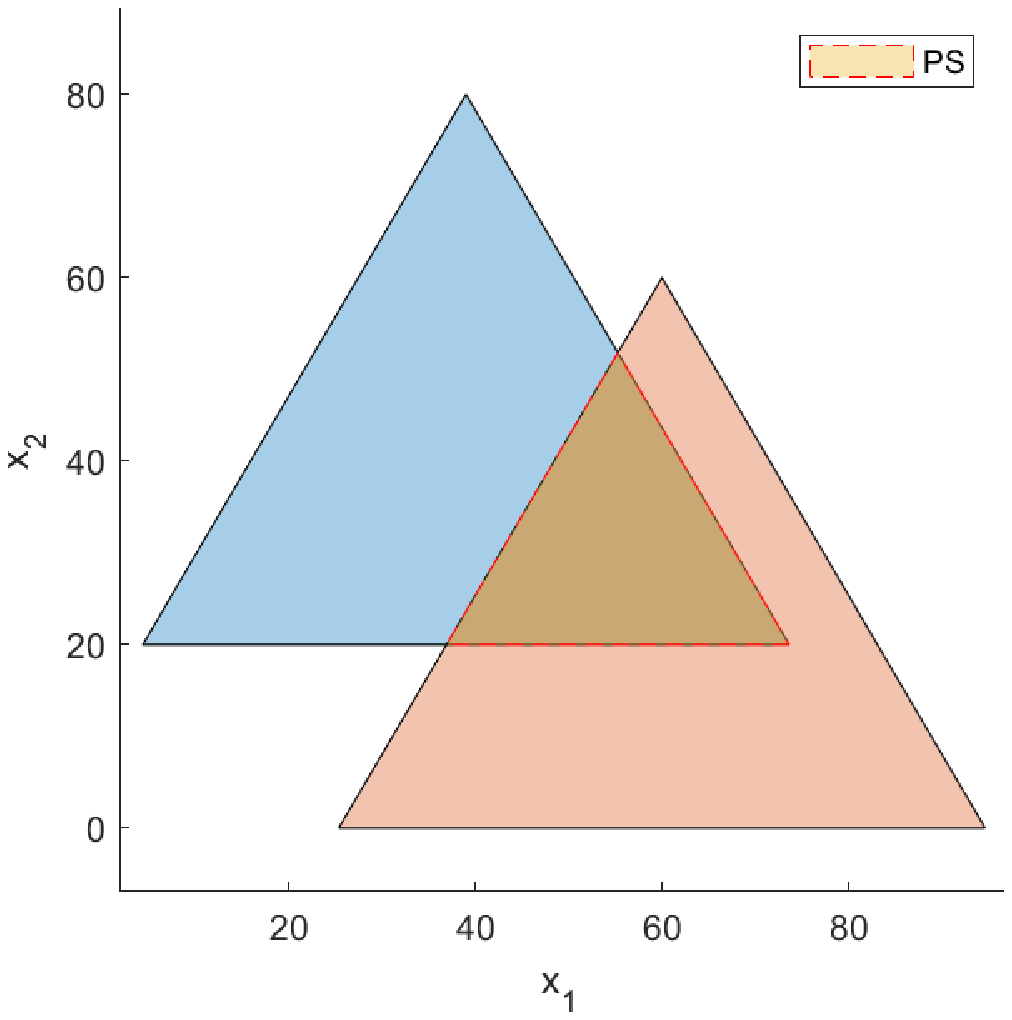}
	\caption{MPDMP5, each triangle represents the PS of one party}
	\label{figMPDMP5}
\end{figure}

\begin{figure}
		\centering
		\includegraphics[width=.8\textwidth]{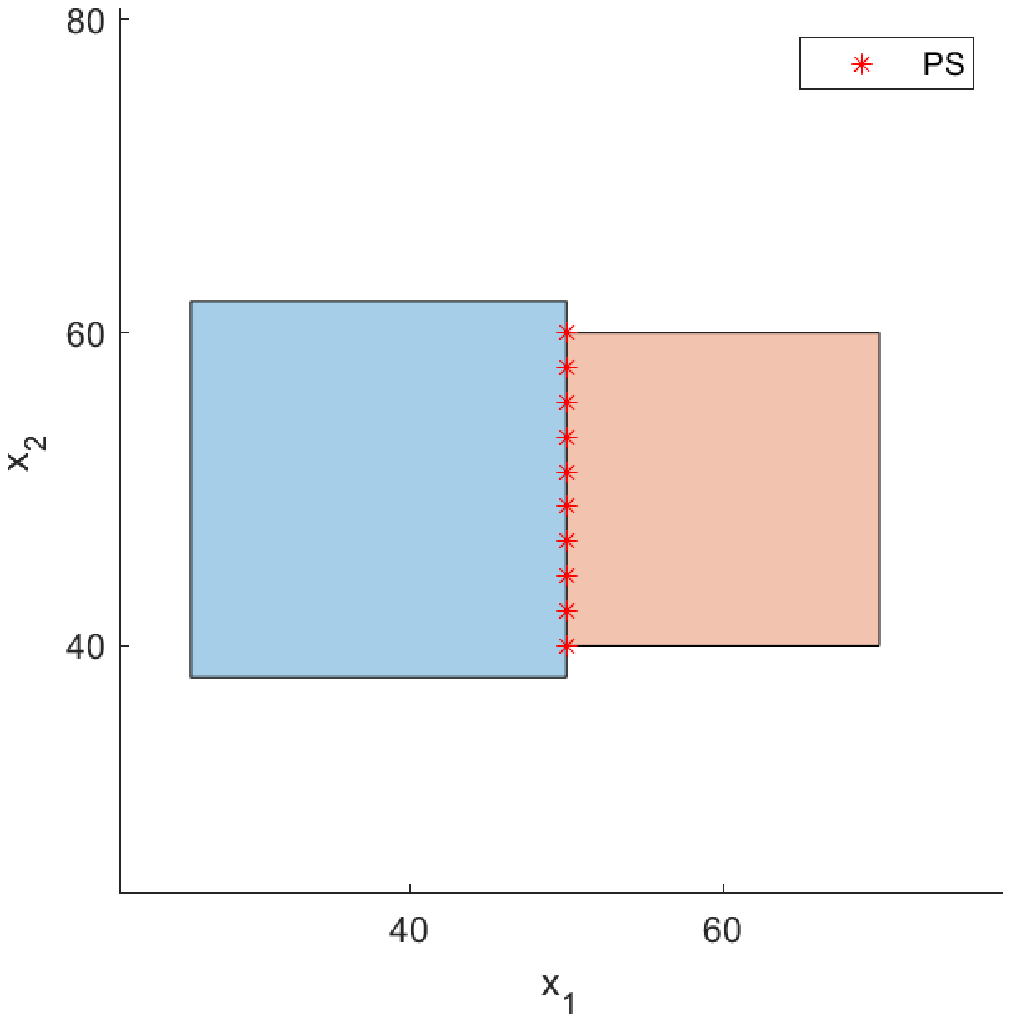}
		\caption{MPDMP6, each rectangle represents the PS of one party}
		\label{figMPDMP6}
\end{figure}
\begin{figure}
	\centering
	\includegraphics[width=.8\textwidth]{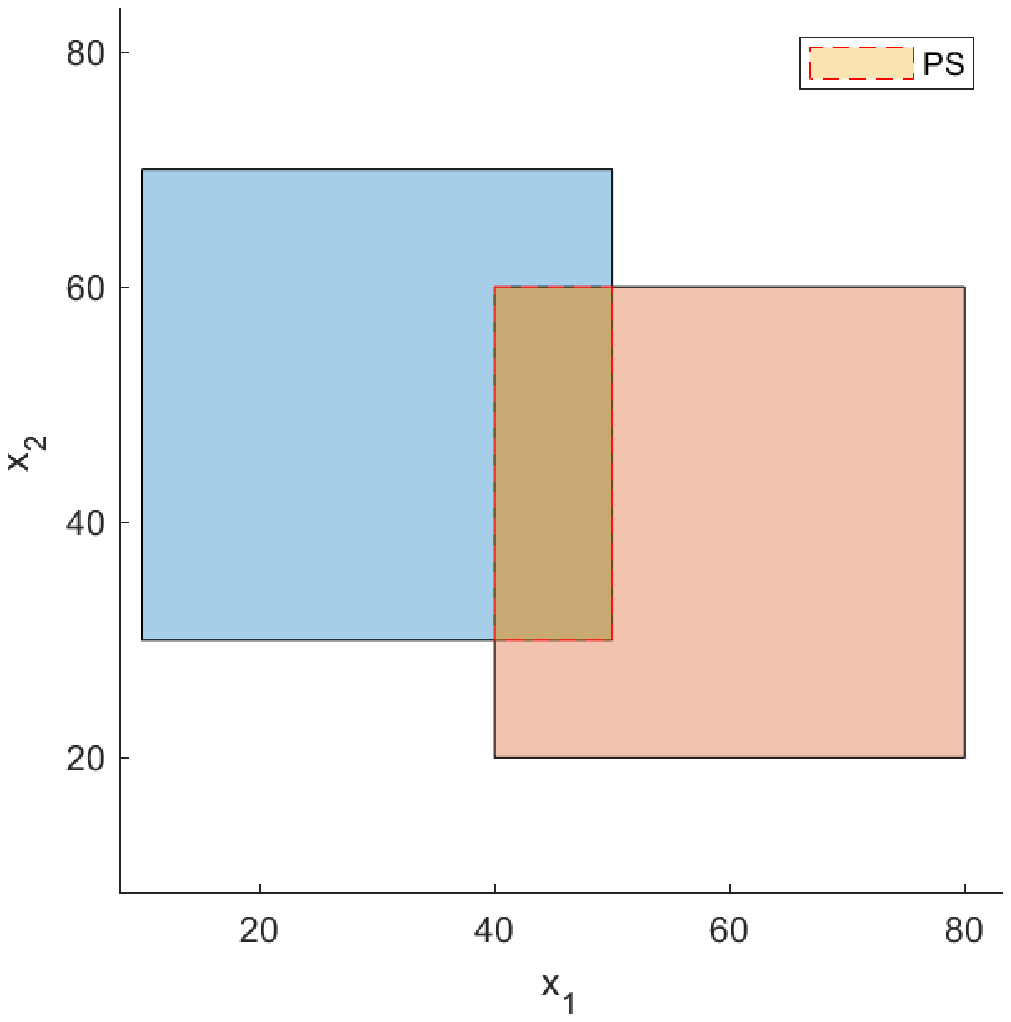}
	\caption{MPDMP7, each rectangle represents the PS of one party}
		\label{figMPDMP7}
\end{figure}
	\begin{figure}
		\centering
		\includegraphics[width=.8\textwidth]{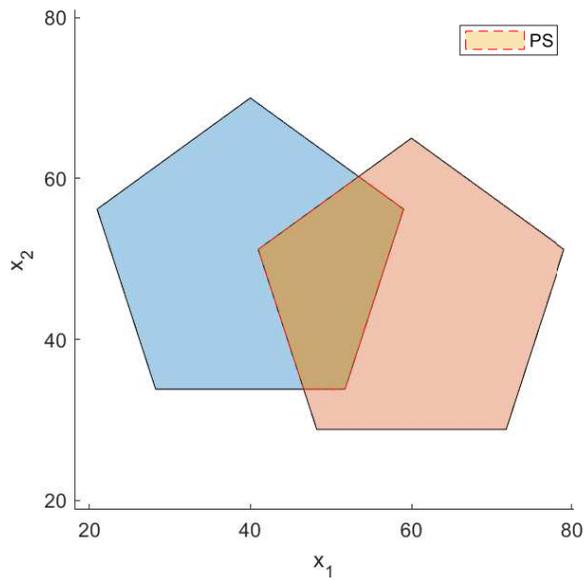}
		\caption{MPDMP8, each pentagon represents the PS of one party}
		\label{figMPDMP8}
\end{figure}

\section{Proposed algorithms}\label{sec:proposed-algorithm}
\subsection{Framework}

\begin{figure}
	\begin{algorithm}[H]
		\caption{OptMPNDS3}
		\label{alg:OptMPNDS3}
		\begin{algorithmic}[1]
			\REQUIRE $N$ (the population size), \\
			$F=(F_1(x),F_2(x),\dots,F_M(x))$ (the objective function) \\
			\ENSURE $MPS$ (the multiparty Pareto optimal solutions)
			\STATE $P_0 = \text{Initialization}(N, F)$
			\STATE $t = 0,Q_t = \emptyset,A = \emptyset$
			\STATE $P_0 = \text{MPNDS2}(P_0, F)$
			\STATE Sort $P_0$ by crowding entropy for each rank
			\WHILE {$FE$ is not reached} \label{step:OptMPNDS3loop1s}
			\STATE Offsprings $Q_t$ are generated from $P_t$ combined with $A$ using parameters $\mu_{CR}$ and $\mu_{F}$
			\STATE The non-dominate solutions in $Q_t$ and $P_t$ are storied in $R_t$, others are put into $A$
			\STATE $B = \text{MPNDS2}(R_t, F)$
			\STATE Sort $B$ by crowding entropy for each rank
			\STATE $i = 1, P_{t+1} = \emptyset$
			\WHILE{$\lvert P_{t+1} \cup B_i \rvert \leq N$}
			\STATE $P_{t+1} = P_{t+1} \cup B_i$
			\STATE $i = i+1$
			\ENDWHILE
				\WHILE{$\lvert P_{t+1} \rvert < N$}
				\STATE Remove the least crowded solution from $B_i$ into $P_{t+1}$
				\STATE Update the crowding entropy of $B_i$
				\ENDWHILE
			\STATE Update the $A$ with individuals not selected
			\STATE Put $CR$ and $F$ which generate solutions in $P_{t+1}$ into $S_{CR}$ and $S_F$
			\STATE Update the parameters $\mu_{CR}$ and $\mu_F$ with $S_{CR}$ and $S_F$
			\STATE $t = t+1$
			\ENDWHILE \label{step:OptMPNDS3loop1e}
			\STATE Set $MPS$ as solutions which is Pareto optimal in all parties in $P_t$
		\end{algorithmic}
	\end{algorithm}
\end{figure}
\begin{figure}[!t]
	\begin{algorithm}[H]
		\caption{Initialization}
		\label{alg:Initial}
		\begin{algorithmic}[1]
			\REQUIRE $N$ (the population size), \\
			$F=(F_1(x),F_2(x),\dots,F_M(x))$ (the objective function) \\
			\ENSURE $\mathcal{S}$ (the Pareto optimal solutions of each party)
			\STATE $\mathcal{S} = \emptyset$
			\FOR{{$i \in \{1,\cdots ,M\}$}} \label{step:Initialloop1s}
			\STATE Initialize population $P_0$ with size $\lfloor \frac{N}{M} \rfloor$
			\STATE $t = 0,Q_t = \emptyset$
			\WHILE {$FEI$ is not reached} 
			\STATE $R_t = P_t \cup Q_t$
			\STATE $A = \text{NonDominatedSorting}(R_t, F_i)$
			\STATE Sort $A$ by crowding distance for each rank
			\STATE $i = 1, P_{t+1} = \emptyset$
			\WHILE{$\lvert P_{t+1} \cup A_i \rvert \leq N$}
			\STATE $P_{t+1} = P_{t+1} \cup A_i$
			\STATE $i = i+1$
			\ENDWHILE
			\STATE $P_{t+1} = P_{t+1} \cup A_i(1:\lfloor \frac{N}{M} \rfloor - \lvert P_{t+1} \rvert)$
			\STATE Create the offspring $Q_{t+1}$ of $P_{t+1}$
			\STATE $t = t+1$
			\ENDWHILE 
			\STATE Put $P_t$ into $\mathcal{S} $
			\ENDFOR \label{step:Initialloop1e}
		\end{algorithmic}
	\end{algorithm}
\end{figure}
\begin{figure}
	\begin{algorithm}[H]
		\caption{MPNDS2 \cite{she2021new}}
		\label{alg:MPNDS2}
		\begin{algorithmic}[1]
			\REQUIRE $R_t, F=(F_1(x),F_2(x),\dots,F_M(x))$
			\ENSURE $\mathcal{F}$
			\STATE $\mathcal{L} = \emptyset$ ;
			\FOR{$i \in \{1,\cdots ,M\}$}
			\STATE $\mathcal{L}(:,i) = \text{NonDominatedSorting}(R_t,F_i)$ ;
			\ENDFOR
			\STATE $\mathcal{F} = \text{NonDominatedSorting}(R_t, \mathcal{L})$
		\end{algorithmic}
	\end{algorithm}
\end{figure}

When it comes to designing an evolutionary algorithm to address MOPs, it is of the utmost importance to balance convergence improvement and diversity maintenance.
That the non-dominated sorting with all objectives together will obtain a lot of dominated solutions for MPMOPs suggests that the selection pressure is so weak, and the diversity of solutions is maintained strongly.
While, the multiparty sorting in OptMPNDS2 that we proposed in \cite{she2021new} generates a relatively strong selection pressure to eliminate the dominated solutions, which results in convergence improving of solutions.

The new algorithm, named OptMPNDS3, employs two strategies including initializing and JADE2 operators.
Algorithm \ref{alg:OptMPNDS3} shows the main pseudocodes of the framework and its main procedures are as follows.
\begin{enumerate}[(1)]
	\item The population $P_0$ is initialized with population size $N$.
	\item \label{step:loop} The trial vectors are generated according to Formula \eqref{eq:DE} and then the crossover is performed.
	      The non-dominated solutions from the population $P_t$ and its trial vectors are gathered into $R_t$;
	      The remaining dominated solutions are storied into archive $A$.
	      These steps are almost the same as JADE2.
	\item Function MPNDS2(.) is applied to sort these individuals into different ranks $B$.
	      Then, individuals in the same rank are sorted according to the crowding distance.
	\item The individuals in sorted collection $B$ are picked as the next generation $P_t$ one by one.
	      And the solutions excluded by non-dominated sorting are put into $A$.
	\item Update the archive $A$ and the parameter $\mu_{CR}$ and $\mu_F$.
	\item Back to \eqref{step:loop} perform the evolution of next generation and the loop from steps \ref{step:OptMPNDS3loop1s} to \ref{step:OptMPNDS3loop1e} is repeated until $FE$ (the maximum fitness evaluations) is reached.
	\item Return the final solutions $MPS$.
\end{enumerate}

A modified JADE2 operator is used to generate the offsprings.
First, when the archive meets the size limitation, the newcomer takes the place of the closest one in archive.
The measure of closeness uses Euclidean distance in the decision space.
Second, crowding distance is substituted by crowding entropy \cite{Wang2010}.
Crowding entropy considers not only the average distance of two points on either side, but also the distribution between the two points.
The source code of this framework is available at https://github.com/MiLabHITSZ/2022SheOptMPNDS3.
\subsection{Initialization}

In the framework, the initializing strategy of OptMPNDS3 is implemented by the function $Initialization(.)$.
And the pseudocodes of function $Initialization(.)$ are shown in Algorithm \ref{alg:Initial}.
The core idea of the initializing is that the solution to the MPMOP has to be Pareto front to all parties if the question has common solutions.
So, the algorithm firstly finds out the Pareto front of each party respectively as the start of the evolution.

Specific, the procedure of function $Initialization(.)$ is depicted as follows.
First, $\mathcal{S}$ is initialized to $\emptyset$.
Then, in the loop from step \ref{step:Initialloop1s} to \ref{step:Initialloop1e}, the Pareto solution of multiple parties are obtained from a party to another.
Algorithm \ref{alg:Initial} treats the MPMOP as multiple MOPs in view of multiple parties.
For each party, the procedure to obtain the Pareto solution is identical to NSGA-II and the maximum fitness evaluations of initialization in each party is set the same as $FEI$.
Finally, the Pareto solutions of all parties, of which quantity is no more than $N$, are gathered into $\mathcal{S}$.

\subsection{Multiparty Sorting}

As in \cite{she2021new}, we first sort the solutions according to non-dominated sorting to obtain the Pareto levels for every party.
And then, the Pareto levels of all parties of one individual are used as its ``objective'' to participate the following sorting to accomplish the evaluation.
The procedure is implemented in function MPNDS2(.), which is depicted in Algorithm \ref{alg:MPNDS2}.
Although MPNDS2(.) is adapted to the framework, its effect is almost the same with \cite{she2021new} as Algorithm \ref{alg:MPNDS2} shown.

\section{Experiments}\label{sec:experiment}

\subsection{Settings}
The test problems are from MPDMP1 from MPDMP8 and the distance function $dis(.)$ of problems uses Euclidean measures.

All these algorithms are run 30 times independently on the problems to compare the average results. For each run, the population size is set to 200; $FE$ (the maximum fitness evaluations) is set to 80000.
In OptMPNDS3, $FEI$, which contains in $FE$, is set as 10000.
For OptAll, OptMPNDS and OptMPNDS2, The simulated binary crossover (SBX) \cite{deb2007self} is used, of which crossover probability is set as 1.0 and the distribution index is set as 15; The polynomial mutation is used, of which the mutation probability is set as 0.5 and the distribution index is set as 20.
For OptMPNDS3, the JADE2 operator is used, where $p$ is set to 0.05 and $c$ is set to 0.05.

\subsection{Performance Metric}
The slightly modified Inverted Generational Distance (IGD), which defines the distance from a point to the PF, is used.
The specific formulas are as follows \cite{liu2020evolutionary}.
\begin{equation}\label{eq:IGD}
	\begin{aligned}
		 & d(v,S)=\min_{s \in S} \sum_{i=1}^{M} \sqrt{(v_{i1}-s_{i1})^2 + \cdots +(v_{ij_i}-s_{ij_i})^2} \\
		 & IGD(P^*,P) = \frac{\sum_{v \in P^*}d(v,P)}{\lvert P^* \rvert}
	\end{aligned}
\end{equation}
where $d(v,S)$ represents the distance from an individual $v$ to the PF $S$.
$s$ represents the individual in $PF$;
$v_{ij} $, as well as $s_{ij}$, means the $j$-th objective of the $i$-th party.
$P^*$ represents the true $PF$ and $P$ represents the $PF$ that algorithms obtained.
The smaller the value is, the better performance the algorithm has.

\subsection{Compared Algorithms}
Algorithm OptAll in \cite{liu2020evolutionary} simply puts objectives of all party together and treats the MPMOP as an MOP.
OptMPNDS and OptMPNDS2 both define the multiparty sorting relationship based on NSGA-II to address the MPMOP.
OptMPNDS takes the max level among the levels obtained by using non-dominated sorting of all parties as the rank of the solution.
Especially, priority is given to the solution with the same level of all parties in OptMPNDS.
OptMPNDS2 directly uses the levels obtained in all parties as the objectives to sort again.
The last procedure of all algorithms is to eliminate the solutions which are not in the first rank after non-dominated sorting in all parties, since all the test problems have the common Pareto optimal solutions.

\subsection{Results}
The IGD values of all these algorithms on problems from MPDMP1 to MPDMP8 are shown in Table \ref{tab0}.
Each row stands for the mean and standard deviation values of certain one algorithm.
And the best results are labeled in the bold font in each column.
Since the DMP is easily used to visually examine the behaviors of multiobjective evolutionary algorithms, this section does not only consider the IGD values, but also analyzes the test algorithms above in the decision space.
\begin{table*}
    \centering
    \caption{The mean and standard IGD values of 30 runs}
    \label{tab0}
    \resizebox{\textwidth}{!}{
        \begin{tabular}{lllll}
            \toprule
            problem	&OptAll	&	OptMPNDS	&	OptMPNDS2	&	OptMPNDS3	\\
            \midrule
            MPDMP1	&4.0422e+00(2.4607e+00)	&	5.4518e-02(2.9469e-02)	&	4.7641e-02(2.7699e-02)	&	\textbf{4.4678e-02}(2.8456e-02)	\\
            MPDMP2	&3.9441e+00(2.0614e+00)	&	6.4854e-02(3.4374e-02)	&	6.1729e-02(3.2529e-02)	&	\textbf{5.7665e-02}(3.5736e-02)	\\
            MPDMP3	&2.7295e+00(6.6283e-01)	&	2.1066e-01(4.7955e-02)	&	2.1530e-01(4.1893e-02)	&	\textbf{1.1286e-01}(3.5309e-02)	\\
            MPDMP4	&1.7758e-02(1.3846e-02)	&	\textbf{9.5788e-04}(2.1092e-03)	&	1.0511e-03(1.6580e-03)	&	1.3263e-02(6.0290e-03)	\\
            MPDMP5	&6.9001e+00(6.0692e-01)	&	2.9676e+00(9.1629e-02)	&	2.9440e+00(1.2337e-01)	&	\textbf{2.6629e+00}(9.8791e-02)	\\
            MPDMP6	&3.2338e+00(7.9582e-01)	&	1.3157e-01(8.9143e-03)	&	1.2986e-01(9.1713e-03)	&	\textbf{8.3832e-02}(1.3690e-03)	\\
            MPDMP7	&8.9266e+00(1.2515e+00)	&	2.4542e+00(1.3254e-01)	&	2.4549e+00(1.3088e-01)	&	\textbf{2.2578e+00}(1.0964e-01)	\\
            MPDMP8	&6.2347e+00(8.0153e-01)	&	2.4560e+00(6.5224e-02)	&	2.4574e+00(8.6273e-02)	&	\textbf{2.2545e+00}(5.5708e-02)	\\
            \bottomrule
        \end{tabular}
    }

\end{table*}

As Table \ref{tab0} shown, the best IGD values of MPDMP1 and MPDMP2 are obtained by the algorithm OptMPNDS3.
These two problems, of which PSs both are a point respectively, show the convergence ability of the OptMPNDS3.
OptMPNDS outperforms other algorithms in MPDMP3, of which PS is a line, and solutions OptMPNDS3 obtained rank align in the PS line.
The performance of OptMPNDS3, especially compared with OptMPNDS, is not so good in the MPMOP4.
In MPMOP5, OptMPNDS3 performs better and finds the optimal region, of which shape is a triangle.
In MPDMP6, MPMOP7 and MPMOP8, where the quantity of objectives of one party are more than three, the solutions OptMPNDS3 obtained almost in the common PS.

\begin{figure}
	\centering
	\includegraphics[width=.8\textwidth]{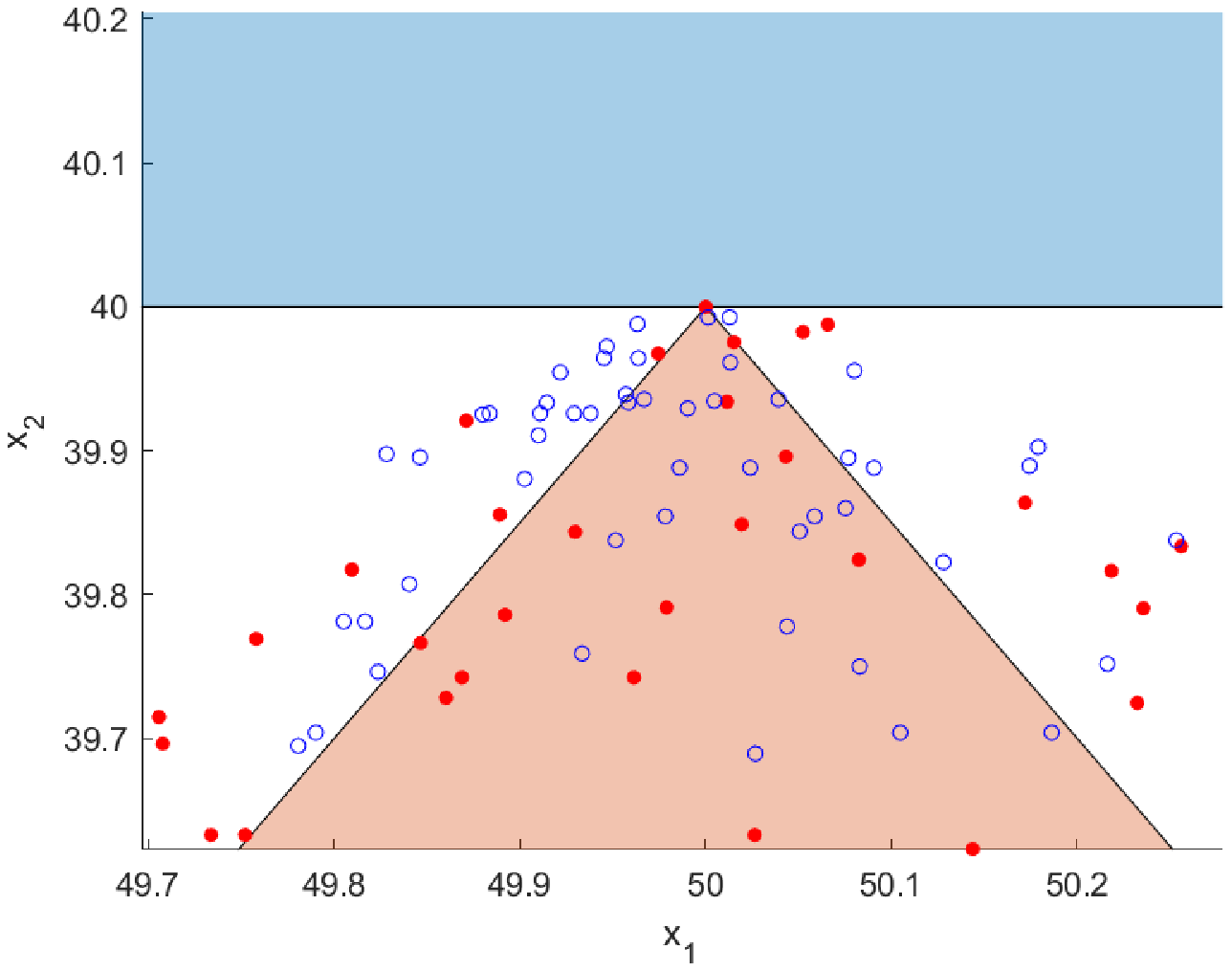}
	\caption{Solutions obtained by one run of OptMPNDS and OptMPNDS3 on MPDMP4}
	\label{fig:problem4}
\end{figure}

\begin{figure}
	\centering
	\includegraphics[width=.8\textwidth]{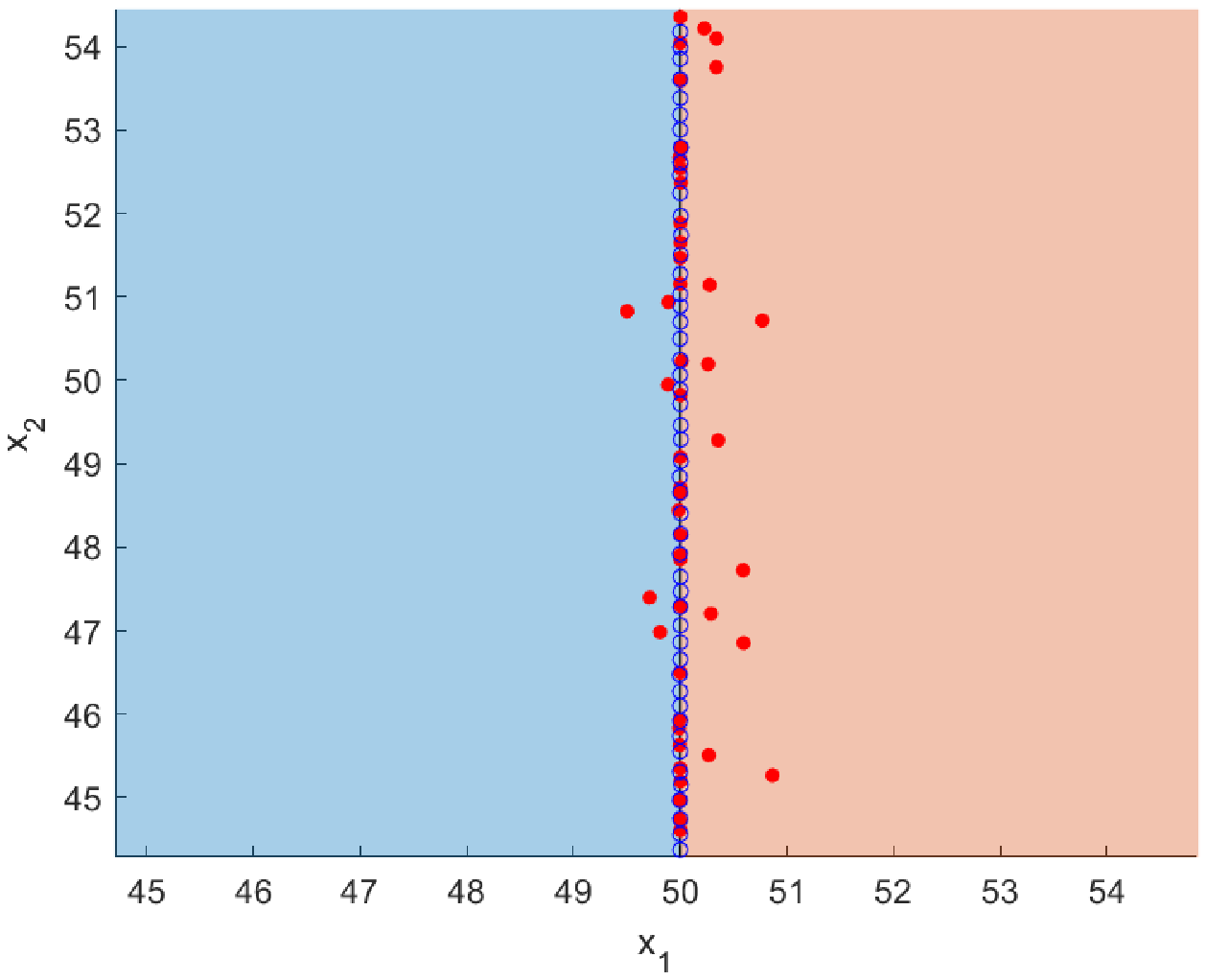}
	\caption{Solutions obtained by one run of OptMPNDS and OptMPNDS3 on MPDMP6}
	\label{fig:problem6}
\end{figure}

The performance of OptMPNDS3 is noticeably different with other algorithms in MPDMP4 and MPDMP6.
The optimal solutions in the final population of one run for the two algorithms on MPDMP4 and MPDMP6 are displayed in the decision space as Figure \ref{fig:problem4} shown.
The red filled circles represent the solutions obtained by OptMPNDS and the blue circles represent the solutions obtained by OptMPNDS3.
Solutions of OptMPNDS3 almost tightly gather around one point which is the true PS of MPDMP4;
However, solutions of OptMPNDS are distributed dispersedly in a larger region.
To our surprise, one of solutions which OptMPNDS obtained almost hits on the optimal point.
As Figure \ref{fig:problem6} shown, OptMPNDS, which is similar to OptMPNDS3, finds solutions around one line that is the true PS, but sometimes its solutions can reach other region way far from the PS;
The behavior of OptMPNDS3 matches our expectation for it shows a perfect balance of convergence and diversity, whose solutions are ranked on the line of true PS almost with the same interval.
It is believed that the solution of OptMPNDS3 distributes more balance.
In conclusion, OptMPNDS3 performs better other algorithms in almost all the problems.

\section{Conclusion}\label{sec:conclusion}

In this paper, it is the first time that a piece of work shows the applications of MPMOPs.
The multiparty test problems based on distance minimization problems can clearly show the behaviors of the test algorithms in the decision space.
Then, we modify the evolutionary algorithms including using initializing method and self-adaptive DE operators to address these problems.
And the proposed algorithm OptMPNDS3 has a comparable performance against the other algorithms.
For the visual of the problem suite, behaviors of algorithms are analyzed to illustrate how algorithms work and why OptMPNDS3 has a comparable performance.

In the future, we will construct multiparty optimization problems without the common Pareto solution and find out a proper performance metric to fairly measure the solutions algorithms obtained for MPMOPs without the common solution.
We will also apply more evolutionary methods to study MPMOPs and propose more effective evolutionary algorithms  to address MPMOPs.

\section* {Acknowledgment}

This study is supported by the National Natural Science Foundation of China (Grant No. 61573327), and the Guangdong Provincial Key Laboratory (Grant No. 2020B121201001).





\bibliographystyle{model2-names}
\bibliography{ref}







\end{document}